\documentclass[10pt,twocolumn,letterpaper]{article}

\usepackage[pagenumbers]{cvpr} 
\usepackage{multirow}
\usepackage{cuted}
\usepackage{capt-of}
\usepackage{nth}
\usepackage[ruled,linesnumbered]{algorithm2e}
\usepackage{float}

\definecolor{cvprblue}{rgb}{0.21,0.49,0.74}
\usepackage[pagebackref,breaklinks,colorlinks,allcolors=cvprblue]{hyperref}

\def\paperID{3449} 
\def\confName{CVPR}
\def\confYear{2025}

\title{MVBoost: Boost 3D Reconstruction with Multi-View Refinement}
\author{Xiangyu Liu$^{1,2}$ \qquad Xiaomei Zhang$^{1,2}$ \qquad Zhiyuan Ma$^{3,4}$ \qquad Xiangyu Zhu$^{1,2}$ \qquad Zhen Lei$^{1,2,3,4}$ \\
$^1$State Key Laboratory of Multimodal Artificial Intelligence Systems,\\
Institute of Automation, Chinese Academy of Sciences
\\
$^2$School of Artificial Intelligence, University of Chinese Academy of Sciences\\
$^3$The Hong Kong Polytechnic University \\
$^4$Center for Artificial Intelligence and Robotics, Hong Kong Institute of Science \& Innovation,\\
Chinese Academy of Sciences\\
\url{https://github.com/Piggy-ch/MVBoost}
}
\begin{document}

\maketitle

\begin{strip}
\vspace{-1.5cm}
\centering
\includegraphics[width=1.0\linewidth]{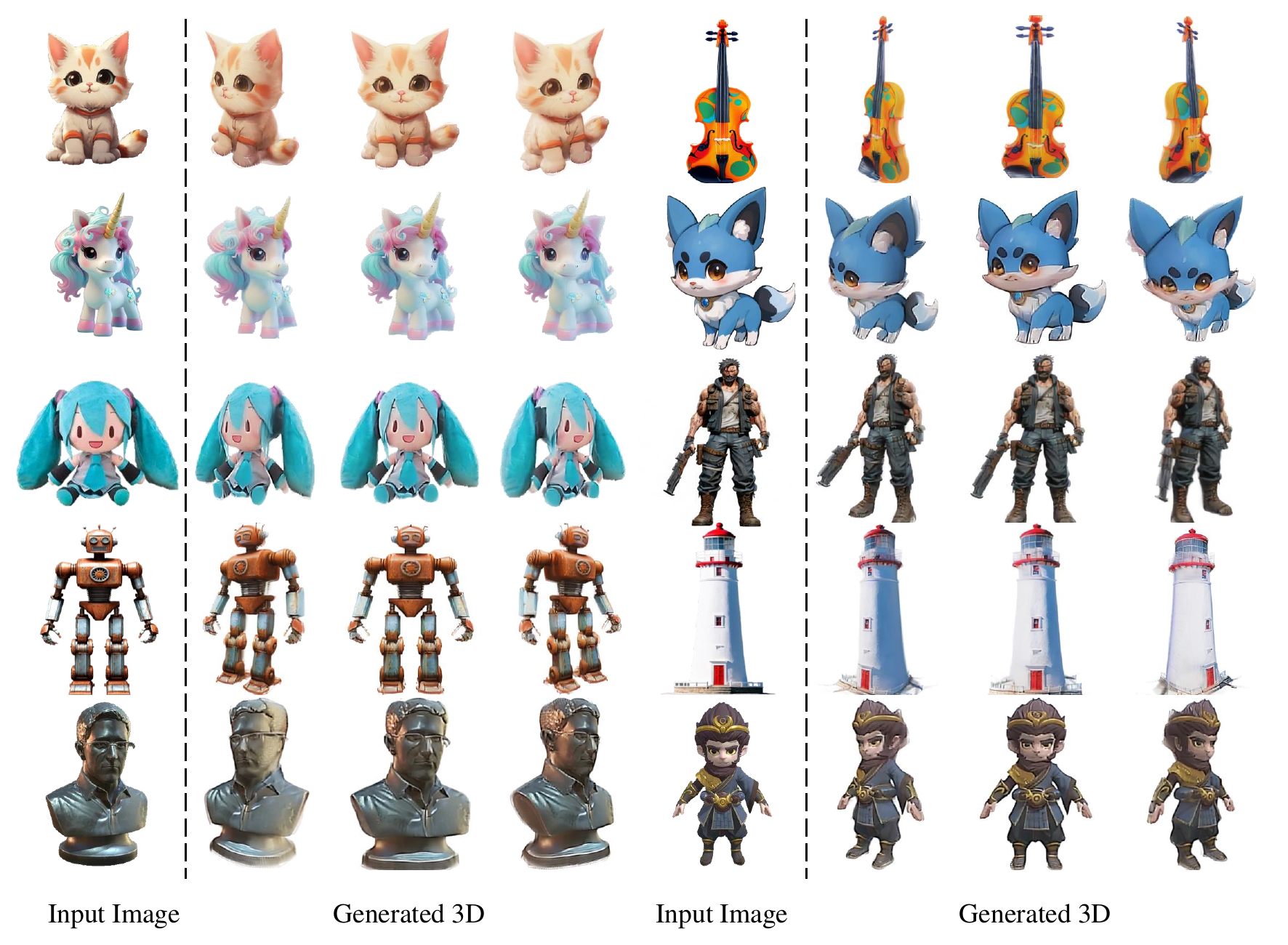}
\captionof{figure}{Given a single image as input, our MVBoost can generate a high-quality 3D asset.}
\label{fig:first}
\end{strip}

\begin{abstract}

Recent advancements in 3D object reconstruction have been remarkable, yet most current 3D models rely heavily on existing 3D datasets. The scarcity of diverse 3D datasets results in limited generalization capabilities of 3D reconstruction models. In this paper, we propose a novel framework for boosting 3D reconstruction with multi-view refinement (MVBoost) by generating pseudo-GT data. The key of MVBoost is combining the advantages of the high accuracy of the multi-view generation model and the consistency of the 3D reconstruction model to create a reliable data source. Specifically, given a single-view input image, we employ a multi-view diffusion model to generate multiple views, followed by a large 3D reconstruction model to produce consistent 3D data. MVBoost then adaptively refines these multi-view images, rendered from the consistent 3D data, to build a large-scale multi-view dataset for training a feed-forward 3D reconstruction model. Additionally, the input view optimization is designed to optimize the corresponding viewpoints based on the user's input image, ensuring that the most important viewpoint is accurately tailored to the user's needs. Extensive evaluations demonstrate that our method achieves superior reconstruction results and robust generalization compared to prior works. 

\end{abstract} 
\section{Introduction}
\label{sec:intro}

Generating 3D assets from a single-view image is a critical task in 3D computer vision~\cite{hong2024lrm,tang2025lgm,wang2024crm,liu2024one2345,liu2023zero,longwonder3d,poole2022dreamfusion}, offering a broad range of applications such as video games~\cite{hao2021gancraft}, virtual reality~\cite{mees2019self}, 3D content creation~\cite{bhatnagar2019multi}, and animation~\cite{liao2023high}. High-fidelity 3D  reconstruction models greatly reduce the labor involved in creating 3D digital assets. However, generating high-fidelity 3D assets from a single-view image while preserving consistent surface details is a challenge, especially for complex objects. A core difficulty in developing these models is the limited availability of high-quality 3D data. Creation of such datasets is a complex task that often requires specialized equipment, advanced capture techniques, or intricate 3D modeling processes. Current publicly available 3D asset datasets~\cite{shapenet2015,deitke2023objaverse,deitke2024objaverse} lack high-quality textures and contain significant repetitiveness, making them insufficient for training 3D generative models effectively.



Several methods leverage 2D diffusion models to enhance 3D generative models due to the widespread success of diffusion models in image generation. DreamFusion proposes Score Distillation Sampling (SDS)~\cite{poole2022dreamfusion}, which distills 3D knowledge from 2D diffusions and inspires the advancement of SDS-based 2D lifting methods~\cite{poole2022dreamfusion,wang2024prolificdreamer,ma2025scaledreamer}. Although SDS-based methods can produce highly realistic visual effects, their optimization-based approaches require several hours to generate a refined 3D asset, making them impractical for 3D content creators. Additionally, SDS-based methods often suffer from poor geometry and inconsistencies, such as the ``Janus" problem. To address these issues, various feed-forward methods~\cite{li2024instantd,hong2024lrm,tang2025lgm,liu2024one2345} have developed that use four views of ground truth as input during training, generating 3D assets through a robust 3D reconstruction network. However, these methods rely on multi-view diffusion models~\cite{liusyncdreamer,wang2023imagedreamimagepromptmultiviewdiffusion} during inference, which may yield inconsistent multi-view outputs. In practical application, the reconstructed model will perform poorly if the training set lacks examples for the current scene. Collecting additional data for new scenarios is costly, making reconstruction very inflexible to apply.

In this work, we propose a novel framework for boosting 3D reconstruction with multi-view refinement (MVBoost) from a single-view image. The method consists of a multi-view refinement strategy and a boosting reconstruction model. The multi-view refinement strategy combines the advantages of the high accuracy of the multi-view generation model and the consistency of the 3D reconstruction model, as our data source. While, the multi-view generation model excels in accuracy but lacks consistency across views, whereas the 3D reconstruction model provides consistent multi-view data, but with lower accuracy. Specifically, given a single-view input image, the multi-view refinement strategy uses a multi-view diffusion model to generate high accuracy multi-views. Then, these multi-view images are sent into a large 3D reconstruction model to produce consistent original 3D representation. By refining multi-view images rendered from original 3D data, the multi-view refinement strategy generates a large-scale multi-view dataset to train a feed-forward 3D reconstruction model. The boosting reconstruction model using the recently proposed Large Multi-View Gaussian Model as our starting point, we introduce LoRA~\cite{hu2022lora} to stabilize the training process. In practical applications, the alignment of the reconstructed 3D model with the input image is one of the key criteria for assessing the quality of the reconstruction. Therefore, we provide an input view optimization process to optimize the corresponding viewpoint image.          


We conduct extensive experiments on the GSO~\cite{downs2022googlescannedobjectshighquality} dataset. Both qualitative and quantitative results demonstrate that our proposed method outperforms related reconstruction methods. In summary, our contributions are:

\begin{itemize}

\item We present a novel framework for constructing pseudo-ground truth, which harnesses the advantages of high-accuracy multi-view generation to improve overall reconstruction quality.

\item We propose a sophisticated framework designed to integrate diverse single-view datasets into the 3D reconstruction training. This framework is further refined based on user-provided input images, culminating in a significant enhancement of the reconstruction results.

\item Results validate that our method could reconstruct high-fidelity 3D reconstruction and achieve new state-of-the-art results.

\end{itemize}


\section{Related Work}
\label{sec:related_work}

\begin{figure*}[t]
  \centering
    \includegraphics[width=1.0\linewidth]{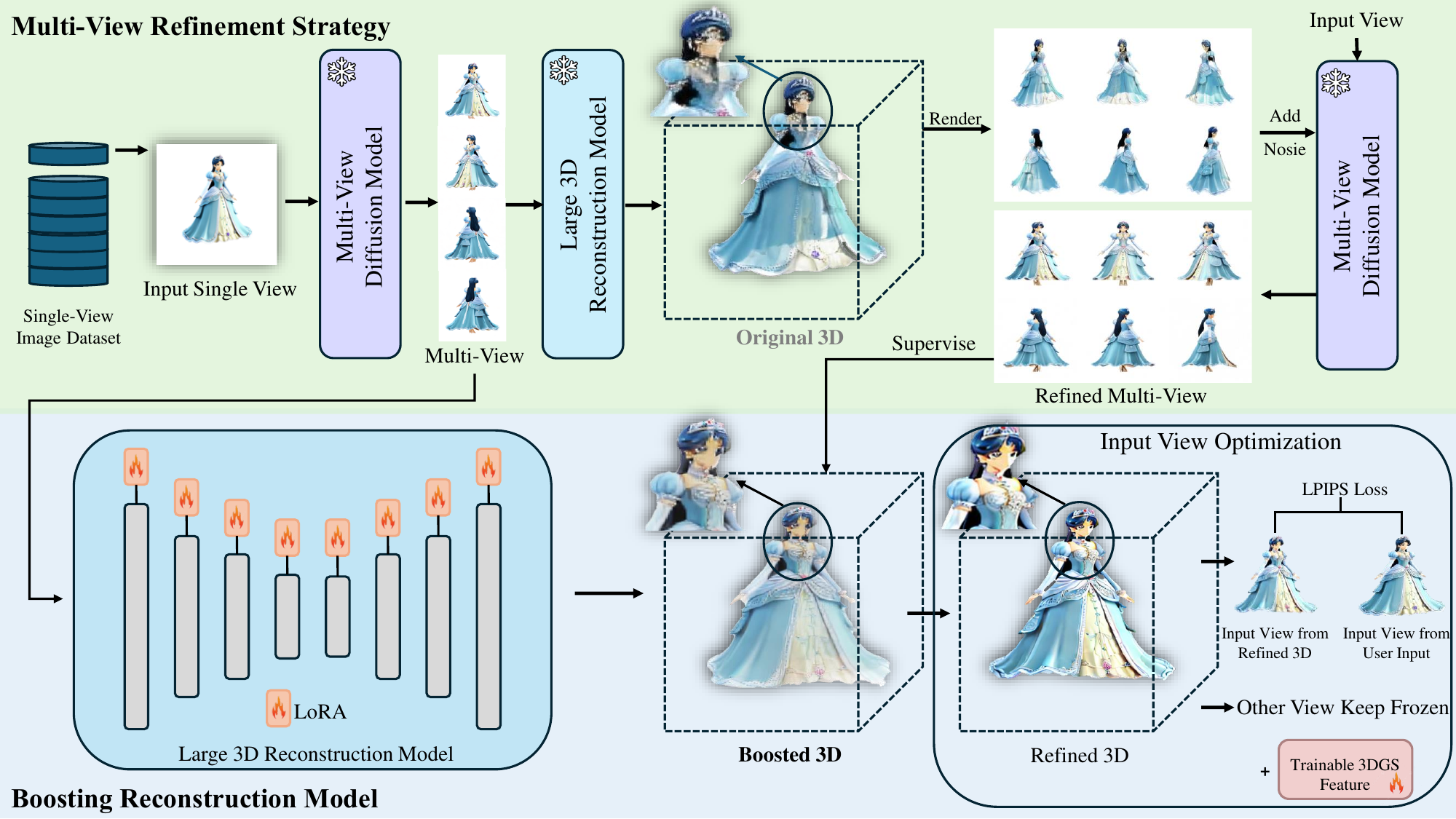}
    \caption{The overview of our MVBoost framework. Given a single-view image dataset, we first employ a multi-view diffusion model to generate the original multi-view dataset. Then the original multi-view is sent into a large 3D reconstruction model to produce the 3D Gaussian Splatting. Several views are rendered from this 3D Gaussian Splatting, and refined by the diffusion model to produce the refined multi-view dataset. During training, the refined multi-view dataset is used to supervise the 3D reconstruction model with LoRA. Finally, the generated 3D assets are optimized to align with the specific input viewpoint, obtaining high-fidelity reconstruction results.}
    \label{fig:pipeline}
\end{figure*}

\noindent\textbf{3D Generation through Score Distillation.} Learning to generate 3D assets from single images presents a formidable challenge due to the limited availability of 3D-image paired data. DreamFusion~\cite{poole2022dreamfusion} introduces Score Distillation Sampling (SDS), along with the concurrent effort Score Jacobian Chaining~\cite{sjc}, which incrementally integrates scores from various perspectives to form a 3D representation. Building on SDS, numerous subsequent studies have emerged. Notably, ProlificDreamer~\cite{wang2024prolificdreamer} aims to minimize the noise prediction error to ensure that the model's output closely aligns with the desired distribution of rendered images. Taking it a step further, ScaleDreamer~\cite{ma2025scaledreamer} presents Asynchronous Score Distillation (ASD), which seeks to minimize the noise prediction error without altering the weights of the pre-trained diffusion network. Nevertheless, due to recent limitations on the number of perspectives, SDS-based optimization methods often encounter the ``Janus'' problem, a common issue where multiple conflicting views lead to inconsistencies in the generated 3D asset.

\noindent\textbf{Multi-view Diffusion Models.}
Due to the limited availability of 3D datasets, many approaches adapt 2D diffusion models for 3D reconstruction. For example,  Zero123~\cite{liu2023zero} demonstrates that Stable Diffusion can be fine-tuned to generate novel views by conditioning on relative camera poses. To address inconsistencies among the generated views in Zero123, several studies~\cite{longwonder3d,li2024era3d,wang2024crm,liusyncdreamer,shi2023zero123plus,liu2023one2345++,wang2023imagedreamimagepromptmultiviewdiffusion} have focused on fine-tuning 2D diffusion models to simultaneously generate multiple consistent views of the same object. Wonder3D~\cite{longwonder3d} uses Cross-Domain Diffusion to produce more consistent RGB and normal images. CRM~\cite{wang2024crm} incorporates the Canonical Coordinates Map (CCM) into multi-view generation to enhance the geometric features of 3D assets. Building on Wonder3D, Era3D~\cite{li2024era3d} employs a regression and conditioning scheme to unify arbitrary camera inputs into canonical camera outputs.

\noindent\textbf{Multi-view to 3D Reconstruction.}
In the field of 3D reconstruction and generation, a significant trend is to directly learn a model that reconstructs 3D representations from multi-view based on 3D datasets. LRM~\cite{hong2024lrm} utilizes a transformer backbone to map multi-view image tokens to an implicit 3D triplane NeRF~\cite{mildenhall2021nerf}. The CRM~\cite{wang2024crm} replaces the transformer backbone with a convolutional U-Net backbone and employs a differentiable Flexicubes~\cite{shen2023flexicubes} for direct optimization on the grid. Taking advantage of the speed and quality improvements offered by 3D Gaussian Splatting, both LGM~\cite{tang2025lgm} and GRM~\cite{xu2024grm} substitute triplane NeRF with 3D Gaussian Splatting~\cite{kerbl3Dgaussians}. The VFusion3D~\cite{han2024vfusion3d} fine-tuning video diffusion model generates multi-view data for the refinement of LRM~\cite{hong2024lrm}. Several other works have also made commendable progress in this feed-forward approach. However, training predominantly on the publicly available 3D dataset Objaverse~\cite{deitke2023objaverse}, and employing input views with domain inconsistencies during training and inference, have resulted in limited practicality and generalization in real-world applications. In contrast, our research incorporates a diverse range of 2D datasets with the multi-view refinement strategy into the model training process, ensuring domain-consistent input views throughout both training and inference to improve the generalization and robustness of the feed-forward model.

\section{Method}
\label{sec:method}

The overall framework of our network, MVBoost, is illustrated in Figure \ref{fig:pipeline}. We first provide an overview of the methods and processes behind diffusion models and 3D generation in Section \ref{sec:Preliminaries}. Next, we present our multi-view refinement strategy (Section \ref{sec:Multi-View Refine Model}), which addresses inconsistencies across views generated by the diffusion model. In Section \ref{sec:Boosting Reconstruction Model}, we describe our boosting reconstruction model, which leverages a refined 2D multi-view dataset to achieve high-fidelity 3D reconstruction without relying on any 3D datasets. Finally, in Section \ref{sec:Input View Optimization}, we explain how we optimize the alignment of the 3D Gaussian Splatting with the corresponding input viewpoint for improved 3D representation.


\subsection{Preliminaries}
\label{sec:Preliminaries}

\noindent\textbf{Diffusion Model.}
Diffusion Model~\cite{ho2020denoising,sohl2015deep,dhariwal2021diffusion} is a generative model that involves forward and reverse diffusion processes. In forward diffusion, data $x_0$ is incrementally corrupted with Gaussian noise ($x_t = {{\alpha}_t}x + \sigma_t\epsilon, t \in \{1,...,T\}$) to approach a noise distribution. The reverse process, which is the generative phase, uses a neural network to predict and apply denoising steps, progressively recovering the original data as it transitions from $t=T$ back to $t=0$.


\noindent\textbf{3D Generation.} Given a single-view image $c$, we utilize the Multi-View Diffusion Model $\mathcal{G}$ to generate multiple views $C^{\pi}=\{c^{\pi_{i}}\}_{i=1}^{n} $. This process can be represented as
\begin{equation}
   C^{\pi}=\{c^{\pi_{i}}\}_{i=1}^{n} =\mathcal{G}(c_{T}^{\pi_{1}},\cdots,c_{T}^{\pi_{n}};c,T),
\end{equation}
where $c_{T}^{\pi_{1}},\cdots,c_{T}^{\pi_{n}}$ are initialized as Gaussian noise, $T$ represents the timestep, and ${\pi}_i$ corresponds to the camera pose for each view.

Using the multi-view data $C^{\pi}$ as input, we apply a large 3D reconstruction model $\mathcal{R}_{\phi}$ to generate the 3D representation $\theta$:

\begin{equation}
   \theta =  \mathcal{R}_{\phi}(C^{\pi}).
\end{equation}

The 3D representation $\theta $ can be either a NeRF~\cite{Nerf} or 3D Gaussian Splatting~\cite{kerbl3Dgaussians}. In our specific implementation, we utilize 3D Gaussian Splatting. We first generate an original 3D Gaussian Splatting denoted as ${\theta}$ using a standard 3D asset generation process. Then, we render $\theta$ from specific viewpoints to obtain multi-view images, represented as $x^{\pi}$. The rendering process can be be represented as 
\begin{equation}
    x^{\pi} = g({\theta}, \pi).
\end{equation}

\subsection{Multi-View Refinement Strategy}
\label{sec:Multi-View Refine Model}
The single-view image dataset can be enhanced using Multi-View Refinement to generate a large, refined multi-view dataset.

Previous methods often relied on limited 3D data as ground truth, which constrained their performance. While VFusion3D~\cite{han2024vfusion3d} leverages video diffusion models for multi-view generation, it lacks explicit view consistency constraints, relying only on implicitly learned multi-view relationships, which can result in geometric inconsistencies. We propose a method that creates pseudo-ground truth by combining 3D multi-view consistency with the high accuracy of the diffusion model. The process is as follows:

In our forward refinement diffusion, the Gaussian Splatting rendered view $x^{\pi}$ is progressively corrupted by Gaussian noise, approaching a noise distribution:
\begin{equation}
    x^{\pi}_t = {{\alpha}_t}x^{\pi} + {\sigma}_t\epsilon.
\end{equation}

The reverse refinement diffusion process predicts the denoising operations using $X^{\pi}_t = \{x^{\pi}_t\}_{i=1}^{n}$ and the input view $c$, resulting in the refined multi-view set $C^{\pi}_{\uparrow} = \{c^{\pi}_{\uparrow}\}_{i=1}^{n}$:

\begin{equation}
    C^{\pi}_{\uparrow}=\mathcal{G}(X^{\pi}_t;c,t),
\end{equation}
where $t = sT$ and $T$ represents the maximum timestep. The parameter $s$ controls the strength of the noise added during the process, ranging from 0 to 1.

\subsection{Boosting Reconstruction Model}
\label{sec:Boosting Reconstruction Model}
As illustrated in Figure \ref{fig:pipeline}, we use the refined multi-view dataset generated from Section \ref{sec:Multi-View Refine Model} to boost the reconstruction model eliminating the need for 3D datasets.

We input the original multi-view data $C^{\pi} = {c^{\pi_i}}_{i=1}^{n}$ into our boosted model to ensure consistency between the training and inference inputs. In our model, we apply LoRA exclusively to the original 3D reconstruction model, $\mathcal{R}{\phi}$, specifically within its Cross-view Self-Attention component, resulting in the boosted model $\mathcal{R}{{\phi}^*}$. By training $\mathcal{R}_{{\phi}^*}$ with the refined multi-view dataset  $C^{\pi}_{\uparrow}$, we obtain the boosted 3D Gaussian Splatting,  ${\theta}^{*}=\mathcal{R}_{{\phi}^*}( C^{\pi})$.

In summary, we use a multi-view diffusion model along with a refinement strategy to convert a single-view image into a multi-view tuple  \((C^{\pi}, C^{\pi}_{\uparrow})\). The training process is defined as follows:
\begin{equation}
    {\phi}^* = \arg \min_{{\phi}} \mathbb{E}_{(C^\pi,C^{\pi}_{\uparrow}),i} \left[\mathcal{L}(g(\mathcal{R}_{{\phi}^*}( C^{\pi}),\pi), C^{\pi}_{\uparrow}) \right],
\end{equation}
where \( \mathcal{L} \) is the loss function that supervises the reconstruction model by comparing the render views \( g(\theta^{*},\pi)\) against the refined multi-view \( C^{\pi}_{\uparrow}\). This ensures consistent improvement across different viewpoints and adaptability to various 3D representations. The loss function \( \mathcal{L} \) is a combination of Mean Squared Error (MSE) and LPIPS loss.

The training framework, including the LoRA-augmented Self-Attention, is compatible with various reconstruction models, supporting different types of 3D representations. Detailed algorithmic illustrations can be found in the supplementary material.

Next, we integrate a set of learnable parameters into the generated 3D Gaussian Splatting. This process involves aligning the user-provided input view with the corresponding view in the 3D Gaussian Splatting through Input View Optimization.

\subsection{Input View Optimization}
\label{sec:Input View Optimization}
To improve the alignment between the generated 3D assets and the input view, we search over the possible camera poses of the 3D assets to find the pose that minimizes the LPIPS loss. Let \( \pi \) represent the set of all possible camera poses. The optimal camera pose \(  \pi_{opt} \) is defined as:

\begin{equation}
 \pi_{opt} = \arg \min_{ \pi} \text{LPIPS}(g(\theta^{*},\pi), c),
\end{equation}
where \( g(\theta^{*},\pi) \) denotes the rendered view from the original 3D Gaussian Splatting \( \theta^{*} \) at camera pose \( \pi \), and \( c \) is the input view.

Subsequently, we enhance the 3D Gaussian Splatting \( \theta^{*} \) by adding  a learnable matrix \( W \):

\begin{equation}
\theta_{\uparrow} = \theta^{*} + W.
\end{equation}

This produces the refined 3D Gaussian Splatting \( \theta_{\uparrow} \). The  matrix \( W \) is optimized by minimizing the LPIPS loss between the rendered view at the optimal camera pose \( \pi_{opt}\) and the input view \( c \):

\begin{equation}
W = \arg \min_W \text{LPIPS}(g(\theta_{\uparrow},\pi_{opt}), c).
\end{equation}

During this optimization, only the view corresponding to \( \pi_{opt} \) is optimized, while other views remain frozen. This ensures that the refined 3D Gaussian Splatting \( \theta_{\uparrow} \) aligns more closely with the input view, without affecting the fidelity of the other perspectives.

\section{Experiment}
\label{sec:Experiment}

\begin{figure*}
  \centering
    \includegraphics[width=0.97\linewidth]{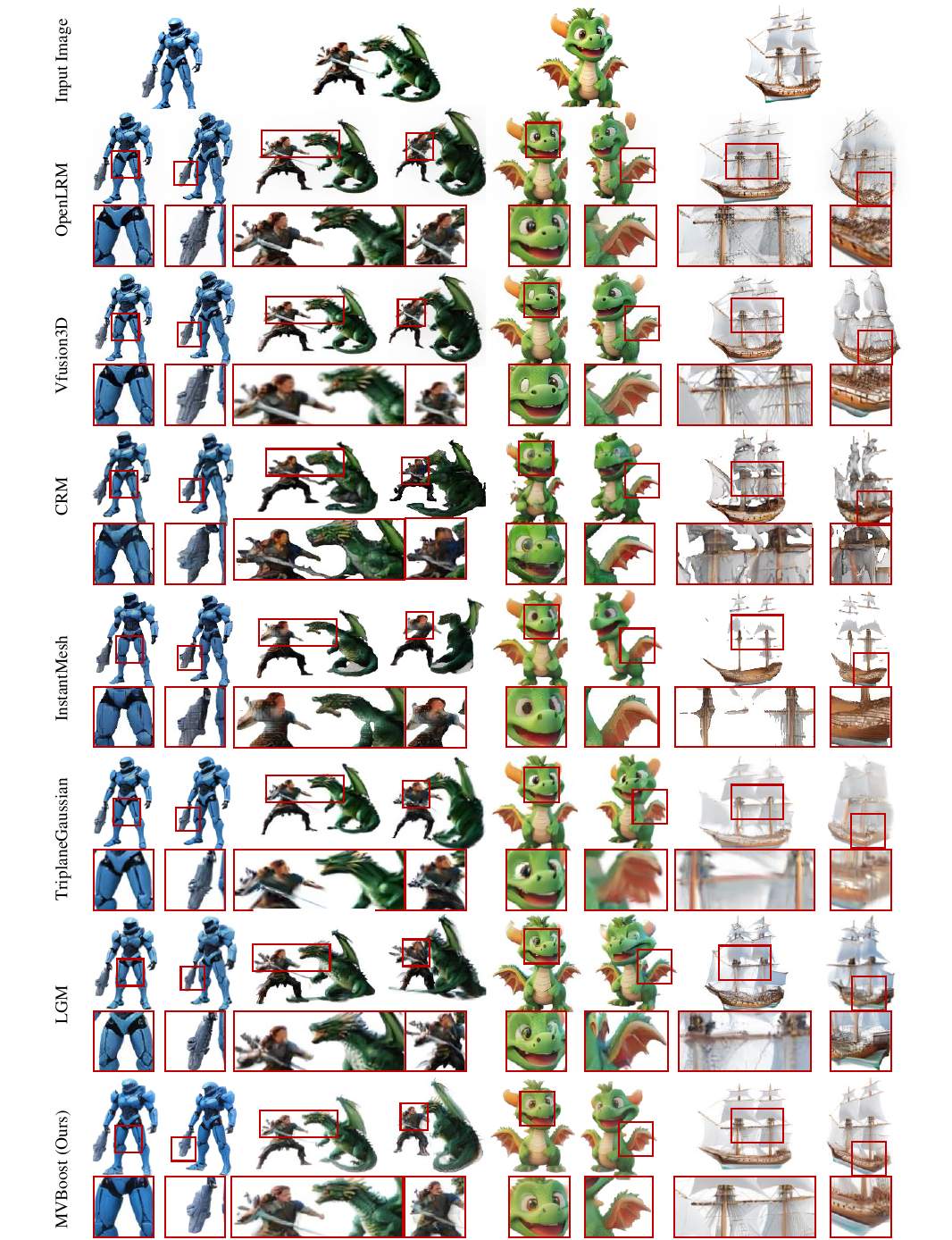}
    \caption{Qualitative comparison of image-to-3D methods. Our approach demonstrates superior 3D generation across a range of challenging images.}
    \label{fig:compare}
\end{figure*}

\subsection{Implementation Details}
\noindent\textbf{Datasets.} We employ ChatGPT to generate over $100k$ prompts for distinct objects in bulk, and utilize a text-to-image model to transform the prompts into finely crafted object images. We do not rely on any existing image datasets for training, which renders our dataset acquisition nearly cost-free, allowing users to expand the image dataset according to their specific needs. Our sota experiments are conducted on the entire Google Scanned Objects (GSO~\cite{downs2022googlescannedobjectshighquality}) dataset, while our ablation studies are carried out on a randomly selected subset of 300 assets from the GSO dataset.

\noindent\textbf{Baselines.} In our research, the Multi-View Diffusion Model utilizes Era3D~\cite{li2024era3d}, while the large 3D reconstruction model employs LGM~\cite{tang2025lgm} which uses Gaussian Splatting for 3D representation. Our pipeline theoretically does not impose any restrictions on the specific types of multi-view diffusion models, reconstruction models, or 3D representations. Among the models we compare, TriplaneGaussian and LGM both use 3D Gaussian Splatting, while OpenLRM~\cite{hong2024lrm} and VFusion3D~\cite{han2024vfusion3d} employ NeRF, and InstantMesh~\cite{xu2024instantmesh} and CRM~\cite{wang2024crm} utilize mesh representations. To ensure fairness, in the main results (Section \ref{sec:main result}), the comparative experiments of our method against other models are conducted without utilizing Input View Optimization.

\noindent\textbf{Training.} We train our model using 8 NVIDIA A100 (80G) GPUs for about a day, which, to our knowledge, represents the lowest training cost and shortest training time compared to similar models~\cite{han2024vfusion3d}. Similar to the training of LGM, we resize the images to 256×256 for LPIPS loss to save memory. Both our training and inference input views come from the unrefined multi-view diffusion. All rendered images and supervision images were set against a white background with a resolution of 512×512.



\subsection{Main Results}
\label{sec:main result}
\noindent\textbf{Quantitative Results.} 
We compare MVBoost with recent open-world feed-forward single/sparse-view to 3D methods, including TriplaneGaussian~\cite{zou2023triplane}, LGM~\cite{tang2025lgm}, OpenLRM~\cite{hong2024lrm}, CRM~\cite{wang2024crm}, and InstantMesh~\cite{xu2024instantmesh}. For the comparative methods, we utilize their official implementations. Since input settings differ among the baselines, we evaluate all methods in a unified single-view to 3D setting. For the GSO dataset, we utilize the first thumbnail image as the single-view input. Then, we use the official multi-view diffusion models of each respective model to generate multi-view inputs, thereby simulating the real user environment.

\begin{table}[!ht]
    \caption{Quantitative comparison for the visual quality on Google Scanned Objects (GSO) orbiting views between our method and baselines for single image to 3D methods}
    \label{Table:2d main}
    \centering
    \begin{tabular}{|l|c|c|c|}
    \hline
        Method & PSNR↑ & SSIM↑ & LPIPS↓ \\ \hline
        OpenLRM~\cite{hong2024lrm} & 16.728 & 0.785 & 0.208  \\ 
        VFsuion3D~\cite{han2024vfusion3d} & 17.416 & 0.846 & 0.155  \\
        CRM~\cite{wang2024crm} & 17.435 & 0.800 & 0.195  \\ 
        InstantMesh~\cite{xu2024instantmesh} & 16.796 & 0.786 & 0.207  \\ \hline
        TriplaneGaussian~\cite{zou2023triplane} & 13.702 & 0.805 & 0.243  \\ 
        LGM~\cite{tang2025lgm} & 17.148 & 0.776 & 0.220  \\ \hline
        MVBoost~(Ours) & \textbf{18.561} & \textbf{0.859} & \textbf{0.131}  \\ \hline
    \end{tabular}
\end{table}

\begin{table}[!ht]
    \centering
    \caption{Quantitative comparison for the geometry quality on Google Scanned Objects (GSO) between our method and baselines for single image to 3D methods}
     \label{Table:geometry main}
    \begin{tabular}{|l|c|c|}
    \hline
        Method & CD↓ & F-Score↑ \\ \hline
        OpenLRM~\cite{hong2024lrm} & 0.14786 & 0.6562 \\ 
        VFsuion3D~\cite{han2024vfusion3d} & 0.16118 & 0.6365 \\
        CRM~\cite{wang2024crm} & 0.12379 & 0.7306 \\ 
        InstantMesh~\cite{xu2024instantmesh} & 0.12334 & 0.7369 \\ \hline
        TriplaneGaussian~\cite{zou2023triplane} & 0.12132 & 0.7598 \\ 
        LGM~\cite{tang2025lgm} & 0.17032 & 0.6349 \\ \hline
        MVBoost~(Ours) & \bf{0.10110} & \bf{0.7977} \\ \hline
    \end{tabular}
\end{table}

\begin{figure*}
  \centering
    \includegraphics[width=1.0\linewidth]{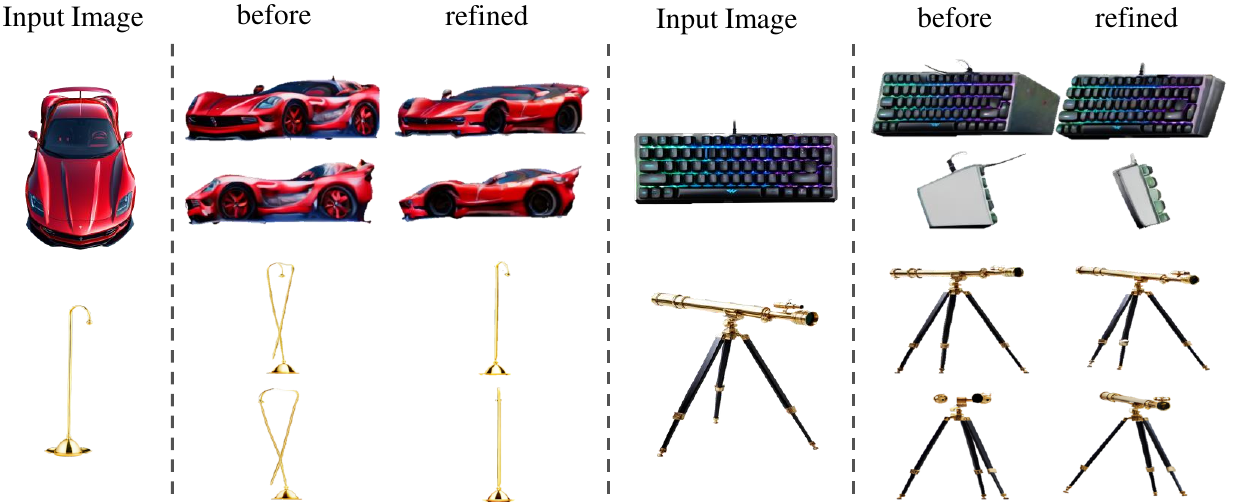}
    \caption{Qualitative comparison of 2D multi-view data before and after refinement. Our refined multi-view shows enhanced geometric correction and superior fidelity in detail reproduction compared to the original.}
    \label{fig:refine}
\end{figure*}

\begin{figure*}
  \centering
    \includegraphics[width=1.0\linewidth]{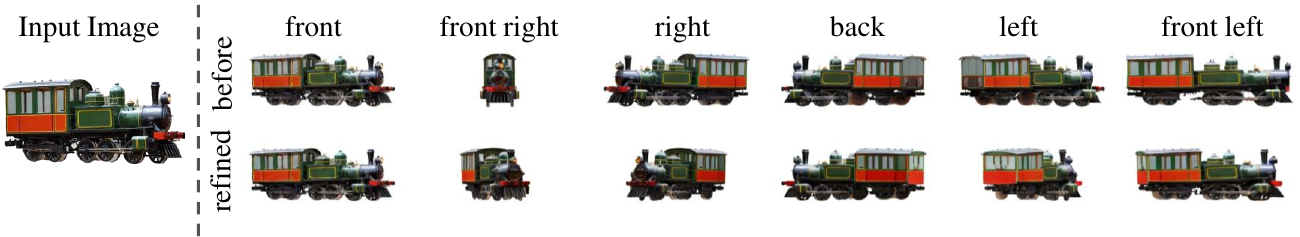}
    \caption{Our multi-view refinement strategy  effectively corrects substantial viewpoint errors in the original views.}
    \label{fig:train}
\end{figure*}

The quantitative experiments on the quality of 2D images are illustrated in Table \ref{Table:2d main}. In this study, our method, TriplaneGaussian, and LGM produce 3D representations using 3D Gaussian Splatting, while OpenLRM and VFusion3D use NeRF, and InstantMesh and CRM employ meshes as their 3D representation. We render the corresponding orbiting views of the 3D Gaussian Splatting, NeRF, and mesh outputs, and compare them against the Ground Truth for evaluation metrics. Our method achieves state-of-the-art performance across various metrics of 2D novel view synthesis quality.

The quantitative experiments on the quality of 3D geometry are illustrated in Table \ref{Table:geometry main}. In this study, our method, TriplaneGaussian, and LGM convert 3D Gaussian Splatting into a mesh following the mesh extraction framework established by LGM. Similarly, OpenLRM and VFusion3D export their NeRF representations as meshes according to their official implementations. Subsequently, all methods are uniformly rescaled to operate at the same mesh scale for the experiments. Even though we choose 3D Gaussian Splatting as our 3D representation, we still achieve state-of-the-art performance across all metrics of geometric quality.

\noindent\textbf{Qualitative Results.} 
Figure \ref{fig:compare} visualizes the qualitative results. Compared with other methods, our boosted results show very high fidelity for diverse inputs, including Robots, cartoon characters, and boat images of various subjects. Our MVBoost correctly models complex geometry (e.g., dinosaur and warrior) and generates the details, such as the ship's sail and halyard, reflecting the great generalization ability of our model.

\subsection{Ablation Study and Disscussion}

\begin{table}[ht]
    \centering
    \caption{The quantitative ablation experiments of the multi-view refinement strategy conducted on the GSO dataset demonstrate that a noise intensity of 0.95 is the most suitable.}
     \label{Table:refine1}
    \begin{tabular}{|c|c|c|c|}
    \hline
        Refine Strength & PSNR↑ & SSIM↑ & LPIPS↓ \\ \hline
        original & 17.811 & 0.815 & 0.118 \\ \hline
        0.10 & 17.798 & 0.818 & 0.114 \\ 
        0.50 & 17.76 & 0.813 & 0.113 \\ 
        0.70 & 17.80 & 0.817 & 0.111 \\ 
        0.90 & 18.27 & 0.821 & 0.106 \\ \hline
        \bf{0.95} & \bf{19.132} & \bf{0.827} & \bf{0.100} \\ \hline
        1.00 & 18.583 & 0.825 & 0.104 \\ \hline
    \end{tabular}
\end{table}

To demonstrate the effectiveness of the multi-view refinement strategy and the input view optimization model, we conduct several ablation studies, which include: (1) qualitative and quantitative comparative experiments on 2D multi-view data before and after refinement; (2) quantitative experiments employing different intensities of refined multi-view supervision; (3) comparative experiments evaluating the Input View Optimization module before and after its implementation.
\begin{figure}[t]
  \centering
  \includegraphics[width=1.0\linewidth]{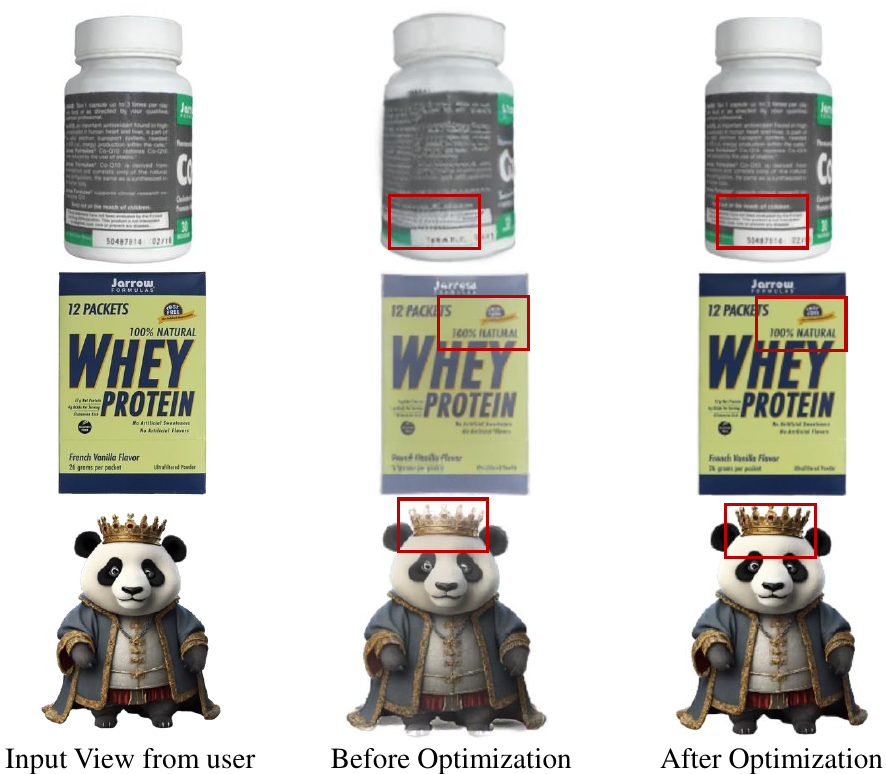}

   \caption{Post-processing with Input View Optimization aligns the 3DGS rendered views with the user's input view.}
   \label{fig:opt}
\end{figure}

Figure \ref{fig:refine} illustrates the qualitative experiments conducted with our multi-view refinement strategy. The first row shows the original multi-view images, while the second row presents the refined multi-view images. The restoration of the red sports car’s side door highlights the module's ability to enhance texture details. The chandelier and telescope examples demonstrate its effectiveness in preserving geometric consistency. In the keyboard example, the side view, which initially appeared unrealistic, is corrected to a more plausible representation. In the train example (Figure \ref{fig:train}), significant viewpoint errors in the original multi-view images are successfully corrected by our method to achieve accurate alignment.

\begin{table}[ht]
    \centering
    \caption{The quantitative ablation experiments on the GSO dataset using a reconstruction model supervised with different noise strengths. }
     \label{Table:refine2}
    \begin{tabular}{|c|c|c|c|}
    \hline
        Refine Strength & PSNR↑ & SSIM↑ & LPIPS↓ \\ \hline
        original & 17.851 & 0.837 & 0.147 \\ \hline
        0.10 & 17.624 & 0.839 & 0.141 \\ 
        0.50 & 17.764 & 0.840 & 0.140 \\ 
        0.70 & 17.904 & 0.840 & 0.139 \\ 
        0.90 & 18.021 & 0.841 & 0.139 \\ \hline
        \bf{0.95} & \bf{18.093} & \bf{0.842} & \bf{0.138} \\ \hline
        1.00 & 18.053 & 0.840 & 0.141 \\ \hline
    \end{tabular}
\end{table}

In the selection of noise intensity within the multi-view refinement strategy, we conduct quantitative experiments on the GSO dataset, as illustrated in Table \ref{Table:refine1}. The results indicate that the optimal refinement effect is achieved at a strength of 0.95. To further illustrate the efficacy of a noise intensity of 0.95, we employ a reconstruction model supervised with varying noise intensities, subsequently conducting quantitative experiments on image quality using the GSO dataset. The experimental results are presented in the Table \ref{Table:refine2}.
D
The input view optimization is a post-processing method. All previous quantitative and qualitative experiments are conducted without Input View Optimization. The experimental results with Input View Optimization are shown in Figure \ref{fig:opt}. On the GSO dataset, the LPIPS loss of the input view decreased from an average of 0.108 to 0.002.

\section{Conclusion}
This paper designs a novel framework MVBoost to boost 3D reconstruction. MVBoost combines the advantages of the high accuracy of the multi-view generation model and the consistency of the 3D reconstruction model to generate pseudo-ground truth, as the data source. A feed-forward 3D reconstruction model is trained by synthetic data, showing superior performance in the reconstruction of 3D assets. Followed by a post-processing method, input-view optimization further optimizes the corresponding viewpoints based on the input image, ensuring that the most important viewpoint is accurately tailored to the user’s needs. Extensive evaluations on benchmark tasks demonstrate that MVBoost achieves state-of-the-art performance in 3D reconstruction. Our approach provides a new pipeline for 3D reconstruction, and we hope to contribute to the development of the 3D reconstruction community.
{
    \small
    \bibliographystyle{ieeenat_fullname}
    \bibliography{main}
}

\clearpage
\setcounter{page}{1}
\maketitlesupplementary
\section{More Implementation Details}

\noindent\textbf{Training.} In our training process, we use a fixed set of six viewpoints (front, front right, right, back, left, front left) for supervision. Our camera model employed orthographic projection. The rank of the LoRA layer in our boosted model is 32.

\noindent\textbf{Metric.}
We evaluate both the 2D visual quality and 3D geometric quality of the generated assets. We use the same single view and then employ each model's official multi-view generation process to create their multi-view inputs, simulating real user inference scenarios. For 3D geometric evaluation, we first align the coordinate system of the generated meshes with the ground truth meshes, and then reposition and re-scale all meshes into a cube of size $[-1, 1]^{3}$. We report Chamfer Distance (CD) and F-Score (FS) with a threshold of 0.05, which are computed by all vertexes from the surface uniformly.

\section{More Details about Method}
\label{sec:Details Method}
MVBoost generates refined multi-view as pseudo-ground truth through the Multi-View Refinement Strategy. The algorithm details are presented in Algorithm \ref{alg:multi-view}. The symbols used in the algorithm are explained and defined in the main paper.
\begin{algorithm}[ht]
    \caption{Multi-View Refinement Strategy}
    \label{alg:multi-view}

    \KwIn{Single-View Image $c$, refine strength $s$}
    \KwOut{Refined Multi-View $C^{\pi}_{\uparrow} = \{c^{\pi_i}_{\uparrow}\}_{i=1}^{n}$}

    $C_T^\pi \sim \mathcal{N}(0,I)$
    
    $ C^{\pi}=\{c^{\pi_{i}}\}_{i=1}^{n} \leftarrow \mathcal{G}(C_{T}^{\pi} ;c,T)$
    
     $\theta \leftarrow   \mathcal{R}_{\phi}(C^{\pi})$

    $X^{\pi}=\{x^{\pi_i}\}_{i=1}^{n} \leftarrow \{g({\theta}, \pi_i)\}_{i=1}^n$
    
    $X_t^\pi=\{x_t^{\pi_i}\}_{i=1}^n\leftarrow\{{{\alpha}_t}x^{\pi_i} + {\sigma}_t\epsilon\}_{i=1}^n$

    $t = sT$

    $C^{\pi}_{\uparrow} = \{c^{\pi_i}_{\uparrow}\}_{i=1}^{n}\leftarrow\mathcal{G}(X^{\pi}_t;c,t)$
\end{algorithm}

We leverage the generated pseudo-ground truth to boost the 3D reconstruction model. LoRA is integrated into the self-attention and cross-attention modules of the model, with its parameters trained using the refined multi-view as supervision. The algorithm details are presented in Algorithm \ref{alg:BoostingRM}. The symbols used in the algorithm are explained and defined in the main paper.








\begin{algorithm}[ht]
\caption{Boosting Reconstruction Model}
\label{alg:BoostingRM}

\KwIn{Multi-View Dataset $\mathbb{S}$, Base Model Parameters $\phi$.}
\KwOut{Optimized model parameters $\phi^*$.}

Freeze the base model parameters $\phi$

Initialize the model trainable parameters $\phi^*$ \;
\While{not converged}{
    Sample a batch $(C^{\pi}, C^{\pi}_{\uparrow})$ from $\mathbb{S}$ 
    
    \quad ${\theta}^* \leftarrow \mathcal{R}_{\phi^*}(C^{\pi})$ 

    \quad $X^{\pi}_{\phi^*} \leftarrow g(\theta^*, \pi)$ 

    \quad $\mathcal{L} = \mathcal{L}(X^{\pi}_{\phi^*}, C^{\pi}_{\uparrow})$ 


    \quad Update $\phi^*$ with $\nabla_{\phi^*}\mathcal{L}$
}
\Return{$\phi^*$}
\end{algorithm}

\section{Experiment on OpenLRM}
\begin{figure*}[ht]
    \centering
    \includegraphics[width=1.0\linewidth]{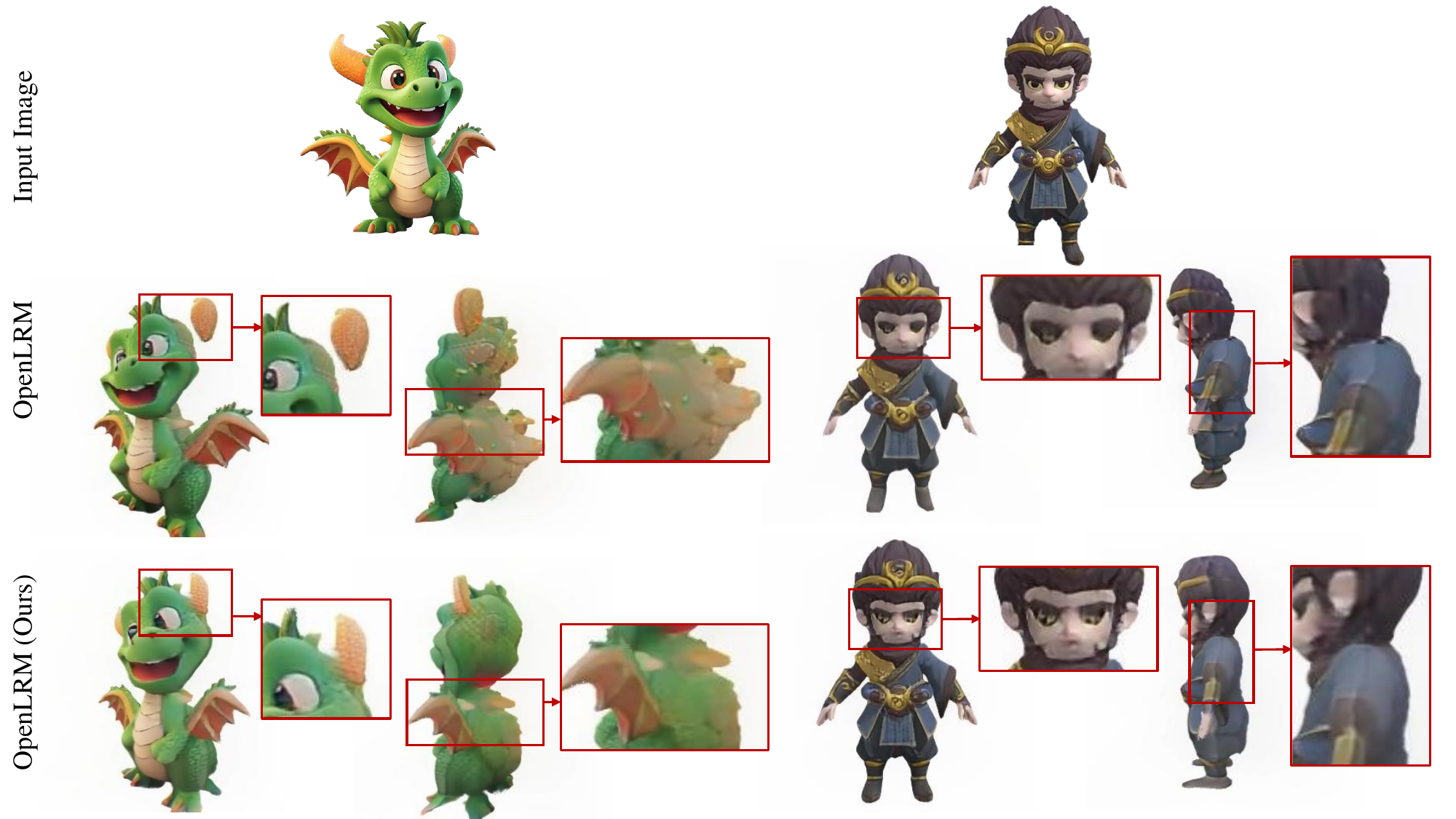}
    \caption{Qualitative experiments of boosted OpenLRM and original OpenLRM.}
    \label{fig:sup}
\end{figure*}
The framework is compatible with various reconstruction models, supporting different types of 3D representations. To further illustrate the versatility of our approach, we boost OpenLRM with multi-view refinement. We employ a dataset of $5k$ refined multi-view images, integrating LoRA layers with the rank of $r=32$ into the self-attention and cross-attention components of the Transformer Decoder in OpenLRM. We train the boosted OpenLRM on 8 NVIDIA A100 (80G) GPUs for half a day. Qualitative results are illustrated in Figure \ref{fig:sup}, while quantitative outcomes are detailed in Table \ref{Table:quan_2d} and Table \ref{Table:quan_geo}. The boost OpenLRM demonstrates superior performance over the original OpenLRM in terms of geometric and textural details.

\begin{table}[!ht]
    \caption{Visual quality comparison on Google Scanned Objects (GSO) between boosted OpenLRM and the original OpenLRM.}
    \label{Table:quan_2d}
    \centering
    \begin{tabular}{|l|c|c|c|}
    \hline
        Method & PSNR↑ & SSIM↑ & LPIPS↓ \\ \hline
        OpenLRM& 16.728 & 0.785 & 0.208  \\ 
        OpenLRM (Ours) & \bf{17.023} & \bf{0.832} & \bf{0.181} \\
        \hline
    \end{tabular}
\end{table}

\begin{table}[!ht]
    \caption{Geometry quality comparison on Google Scanned Objects (GSO) between boosted OpenLRM and the original OpenLRM.}
    \label{Table:quan_geo}
    \centering
    \begin{tabular}{|l|c|c|}
    \hline
        Method & CD↓ & F-Score↑ \\ \hline
        OpenLRM& 0.14786 & 0.6562  \\ 
        OpenLRM (Ours) &\bf{0.12158} & \bf{0.6832} \\
        \hline
    \end{tabular}
\end{table}

\end{document}